\pdfoutput=1

\documentclass[11pt]{article}

\usepackage[]{EMNLP2022}

\usepackage{times}
\usepackage{latexsym}

\usepackage[T1]{fontenc}

\usepackage[utf8]{inputenc}

\usepackage{microtype}

\usepackage{inconsolata}

\usepackage{graphicx}
\usepackage{arydshln}
\usepackage{amsmath}
\usepackage{bm}
\usepackage{booktabs}
\usepackage{pifont}
\usepackage{svg}
\usepackage[colorinlistoftodos, textsize=tiny]{todonotes}
\usepackage{listings}
\definecolor{blued}{RGB}{70,197,221}

\title{Focused Concatenation for Context-Aware Neural Machine Translation}

\author{
    Lorenzo Lupo$^1$ \: Marco Dinarelli$^1$ \: Laurent Besacier$^{2}$ \\
    $^1$Université Grenoble Alpes, France \\
    $^2$Naver Labs Europe, France \\
    \texttt{lorenzo.lupo@univ-grenoble-alpes.fr}\\
    \texttt{marco.dinarelli@univ-grenoble-alpes.fr}\\
    \texttt{laurent.besacier@naverlabs.com}
    }
    
\begin{document}
\maketitle
\begin{abstract}
A straightforward approach to context-aware neural machine translation  consists in feeding the standard encoder-decoder architecture with a window of  consecutive sentences, formed by the current sentence and a number of sentences from its context concatenated to it. In this work, we propose an improved concatenation approach that encourages the model to focus on the translation of the current sentence, discounting the loss generated by target context. We also propose an additional improvement that strengthen the notion of sentence boundaries and of relative sentence distance, facilitating model compliance to the context-discounted objective. We evaluate our approach with both average-translation quality metrics and contrastive test sets for the translation of inter-sentential discourse phenomena, proving its superiority to the vanilla concatenation approach and other sophisticated context-aware systems.
\end{abstract}

\section{Introduction}\label{sec:intro}

While current neural machine translation (NMT) systems have reached close-to-human quality in the translation of decontextualized sentences ~\citep{wu_googles_2016}, they still have a wide margin of improvement ahead when it comes to translating full documents~\citep{laubli_has_2018}. Many works tried to reduce this margin, proposing various approaches to context-aware NMT (CANMT)\footnote{Unless otherwise specified, we refer to \textit{context} as the sentences that precede or follow a \textit{current} sentence to be translated, within the same document.}. A common taxonomy~\citep{kim_when_2019,li_does_2020} divides them in two broad categories: multi-encoding approaches and concatenation (single-encoding) approaches. Despite its simplicity, the concatenation approaches have been shown to achieve competitive or superior performance to more sophisticated, multi-encoding systems~\citep{lopes_document-level_2020,ma_comparison_2021}.
\begin{figure}[]
    \centering
    \includegraphics[width=0.35\textwidth]{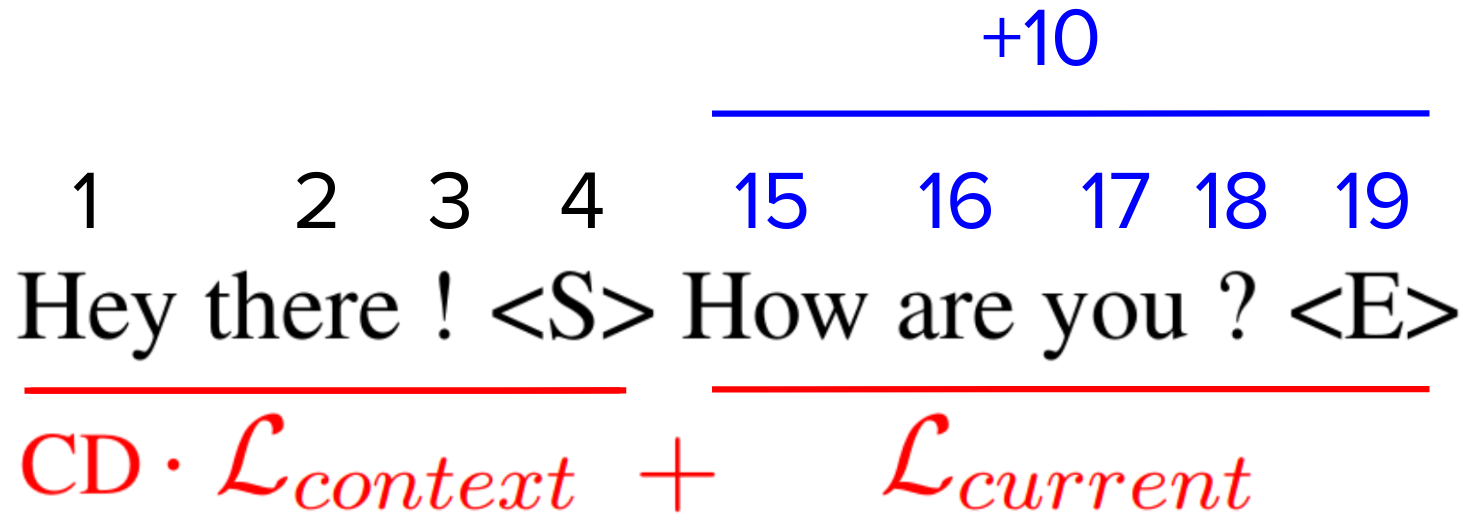}
    \caption{Example of the proposed approach applied over a window of 2 sentences, with context discount {\small CD} and segment-shifted positions by a factor of 10.}
    \label{fig:approach}
\end{figure}
Nonetheless, it has been shown that Transformer-based NMT systems~\citep{vaswani_attention_2017} struggle to learn locality properties~\cite{hardmeier_discourse_2012,rizzi_locality_2013} of both the language itself and the source-target alignment when the input sequence grows in length, as in the case of concatenation \citep{bao_g-transformer_2021}. Unsurprisingly, the presence of context makes learning harder for concatenation models by distracting attention. Moreover, we know from recent literature that NMT systems require context for a sparse set of inter-sentential discourse phenomena only \cite{voita_when_2019,lupo_divide_2022}. Therefore, it is desirable to make concatenation models more focused on local linguistic phenomena, belonging to the current sentence, while also processing its context for enabling inter-sentential contextualization whenever it is needed. We propose an improved concatenation approach to CANMT that is more focused on the translation of the current sentence by means of two simple, parameter-free solutions:
\begin{itemize}
    \item Context-discounting: a simple modification of the NMT loss that improves context-aware translation of a sentence by making the model less distracted by its concatenated context;
    \item Segment-shifted positions: a simple, parameter-free modification of position embeddings, that facilitates the achievement of the context-discounted objective by supporting the learning of locality properties in the document translation task.
\end{itemize}
We support our solutions with extensive experiments, analysis and benchmarking.

\section{Background}\label{sec:background}

\subsection{Multi-encoding approaches}\label{subsec:multienc}

Multi-encoding models couple a self-standing sentence-level NMT system, with parameters $\theta_S$, with additional parameters $\theta_C$ that encode and integrate the context of the current sentence, either on source side, target side, or both. The full context-aware architecture has parameters $\Theta=[\theta_S;\theta_C]$. Multi-encoding models differ from each other in the way they encode the context or integrate its representations with those of the current sentence. For instance, the representations coming from the context encoder can be integrated with the encoding of the current sentence outside the decoder~\citep{maruf_contextual_2018, voita_context-aware_2018,zhang_improving_2018,miculicich_document-level_2018,maruf_selective_2019,zheng_towards_2020} or inside the decoder~\citep{tu_learning_2018, kuang_modeling_2018, bawden_evaluating_2018, voita_when_2019, tan_hierarchical_2019}, by making it attending to the context representations directly, using its internal representation of the decoded history as query.

\subsection{Single-encoder approaches}\label{subsec:singleenc}

The concatenation approaches are the simplest in terms of architecture, as they mainly consist in concatenating each (current) source sentence with its context before feeding it to the standard encoder-decoder architecture~\citep{tiedemann_neural_2017, junczys-dowmunt_microsoft_2019, agrawal_contextual_2018, ma_simple_2020}, without the addition of extra learnable parameters. The decoding can then be limited to the current sentence, although decoding the full target concatenation is more effective thanks to the availability of target context. A typical strategy to train a concatenation approach and generate translations is by sliding windows \citep{tiedemann_neural_2017}. An sKtoK model decodes the translation $\bm{y}_K^j$ of a source window $\bm{x}_K^j$, formed by $K$ consecutive sentences belonging to the same document: the current ($j$th) sentence and $K-1$ sentences concatenated as source-side context. Besides the end-of-sequence token \texttt{<E>}, another special token \texttt{<S>} is introduced to mark sentence boundaries in the concatenation:
\begin{align*}
	\bm{x}_K^j &=\bm{x}^{j-K+1}\text{\tiny <S>}\bm{x}^{j-K+2}\text{\tiny <S>}...\text{\tiny <S>}\bm{x}^{j-1}\text{\tiny <S>}\bm{x}^j \text{\tiny <E>} \\
	\bm{y}_K^j &=\bm{y}^{j-K+1}\text{\tiny <S>}\bm{y}^{j-K+2}\text{\tiny <S>}...\text{\tiny <S>}\bm{y}^{j-1}\text{\tiny <S>}\bm{y}^j \text{\tiny <E>}
\end{align*}
Both past and future contexts can be concatenated to the current pair $\bm{x}^j,\bm{y}^j$, although in this work we consider only the past context, for simplicity. At training time, the loss is calculated over the whole output $\bm{y}_K^j$, but only the translation $\bm{y}^j$ of the current sentence is kept at inference time, while the translation of the context is discarded. Then, the window is slid by one position forward to repeat the process for the $(j+1)$th sentence and its context. Concatenation approaches are trained by optimizing the same objective function as standard NMT over a window of sentences:


\begin{align}\label{eq:concatlikelihood}
	\mathcal{L}(\bm{x}_K^j,\bm{y}_K^j)
	&= \sum_{t=1}^{|\bm{y}^j_K|} \log P(y_{K,t}^j|\bm{y}_{K,<t}^j,\bm{x}_K^j),
\end{align}

so that the likelihood of the current target sentence is conditioned on source and target context.

\subsection{Closing the gap}\label{subsec:closing}

Concatenation approaches have the advantage of treating the task of CANMT in the same way as context-agnostic NMT, which eases learning because the learnable parameters responsible for inter-sentential contextualization are the same that undertake intra-sentential contextualization. Indeed, learning the parameters responsible for inter-sentential contextualization in multi-encoding approaches ($\theta_C$) has been shown to be challenging because the training signal is sparse and the task of retrieving useful context elements difficult \cite{lupo_divide_2022}. Nonetheless, encoding current and context sentences together comes at a cost. In fact, when sequences are long the risk of paying attention to irrelevant elements increases. Paying attention to the "wrong tokens" can harm their intra and inter-sentential contextualization, associating them to the wrong latent features. Indeed, \citet{liu_multilingual_2020} and \citet{sun_rethinking_2022} showed that learning to translate long sequences, comprised of many sentences, fails without the use of large-scale pre-training or data-augmentation (e.g., like \citet{junczys-dowmunt_microsoft_2019} and \citet{ma_comparison_2021} did). \citet{bao_g-transformer_2021} provided some evidence about this leaning difficulty, showing that failed models, i.e., models stuck in local minima with a high validation loss, present a distribution of attention weights that is flatter (with higher entropy), both in the encoder and the decoder, than the distribution occurring in models that converge to lower validation loss. In other words, attention struggles to learn the locality properties of both the language itself and the source-target alignment~\cite{hardmeier_discourse_2012,rizzi_locality_2013}. As a solution, \citet{zhang_long-short_2020} and \citet{bao_g-transformer_2021} propose two slightly different masking methods that allow both the encoding of the current sentence concatenated with context, and the separate encoding of each sentence in window. The representations generated by the two encoding schemes are then integrated together, at the cost of adding extra learnable parameters to the standard Transformer architecture.

\section{Proposed approach}\label{sec:approach}
\subsection{Context discounting}\label{subsec:cd}

Evidently, Equation~\ref{eq:concatlikelihood} defines an objective function that does not factor in the fact that we only care about the translation of the current sentence $\bm{x}^j$, because the context translation will be discarded during inference. Moreover, as discussed above, we need attention to stay focused locally, relying on context only for the disambiguation of relatively sparse inter-sentential discourse phenomena that are ambiguous at sentence level. Hence, we propose to encourage the model to focus on the translation of the current sentence $\bm{x}^j$ by applying a discount $0\leq{\small\text{CD}}<1$ to the loss generated by context tokens:

\begin{align} \label{eq:cd}
	\mathcal{L}_{\small\text{CD}}(\bm{x}_K^j,\bm{y}_K^j)
	&={\small\text{CD}\cdot}\mathcal{L}_{context} + \mathcal{L}_{current}\\
	&={\small\text{CD}\cdot}\mathcal{L}(\bm{x}_{K-1}^{j-1},\bm{y}_{K-1}^{j-1}) + \mathcal{L}(\bm{x}^j,\bm{y}^j). \nonumber
\end{align}

This is equivalent to consider an sKtoK concatenation approach as the result of a multi-task sequence-to-sequence setting~\cite{luong_multi-task_2016}, where an sKto1 model performs the \textit{reference task} of translating the current sentence given a concatenation of its source with K-1 context sentences, while the translation of the context sentences is added as a secondary, complementary task. The reference task is assigned a bigger weight than the secondary task in the multi-task composite loss. As we will see in Section~\ref{subsec:cdanalysis}, this simple modification of the loss allows the model to learn a self-attentive mechanism that is less distracted by noisy context information, thus achieving net improvements in the translation of inter-sentential discourse phenomena occurring in the current sentence (Section~\ref{subsec:main}), and helping concatenation systems to generalize to wider context after training (Section~\ref{subsubsec:robustness}).

\subsection{Segment-shifted positions}

Context discounting pushes the model to discriminate between the current sentence and the context. Such discrimination can be undertaken by cross-referencing the information provided by two elements: sentence separation tokens \texttt{<S>}, and sinusoidal position encodings, as defined in \cite{vaswani_attention_2017}. In order to facilitate this task, we propose to provide the model with extra information about sentence boundaries and their relative distance. 
\citep{devlin_bert_2019} achieve this goal by adding segment embeddings to every token representation in input to the model, on top of token and position embeddings, such that every segment embedding represents the sentence position in the window of sentences. However, we propose an alternative solution  that does not require any extra learnable parameter nor memory allocation: segment-shifted positions. As shown in Figure \ref{fig:approach}, we apply a constant shift after every separation token \texttt{<S>}, so that the resulting token position is equal to its original position plus a total shift depending on the chosen constant \emph{shift} and the index $k={1,2,...,K}$ of the sentence the token belongs to: 
$t'=t+k*shift$. As a result, the position distance between tokens belonging to different sentences is increased. For example, the distance between the first token of the current sentence and the last token of the preceding context sentence increases from $1$ to $1+\textit{shift}$. By increasing the distance between sinusoidal position embeddings\footnote{Positions can be shifted by segment also in the case of learned position embeddings, both absolute and relative. We leave such experiments for future works.} of tokens belonging to different sentences, their dot product, which is at the core of the attention mechanism, becomes smaller, possibly resulting in smaller attention weights. In other words, the resulting attention becomes more localized, as confirmed by the empirical analysis reported in Section~\ref{subsubsec:entropy}. In Section~\ref{subsec:main}, we present results of segment-shifted positions, and then compare them with both sinusoidal segment embeddings and learned segment embeddings in Section~\ref{subsubsec:segemb}.

\section{Experiments}\label{sec:exp}

\subsection{Setup\footnote{See Appendix~\ref{app:setup} for more details.}}\label{subsec:setup}

We conduct experiments with two language pairs and domains. For En$\rightarrow$Ru, we adopt a document-level corpus released by~\citet{voita_when_2019}, based on OpenSubtitles2018 (with dev and test sets), comprised of 1.5M parallel sentences.
For En$\rightarrow$De, we train models on TED talks subtitles released by IWSLT17~\cite{cettolo_wit3_2012}. Models are tested on IWSLT17's test set 2015, while test-sets 2011-2014 are used for development, following related works in the literature.

Besides evaluating average translation quality with BLEU\footnote{Moses' \textit{multi-bleu-detok}~\citep{koehn_moses_2007} for De, \textit{multi-bleu} for lowercased Ru as~\citet{voita_when_2019}.}~\citep{papineni_bleu_2002} and COMET\footnote{Default model: wmt20-comet-da.}~\citep{rei_comet_2020}, we employ two contrastive test suites for the evaluation of the translation of inter-sentential discourse phenomena. For En$\rightarrow$Ru, we adopt \citet{voita_when_2019}'s test suite for evaluation on deixis, lexical cohesion, verb-phrase ellipsis and inflection ellipsis. This test suite is comprised of a development set with examples of deixis and lexical cohesion, that we adopted  for a preliminary analysis of context discounting. For En$\rightarrow$De, we evaluate models on ambiguous pronoun translation with ContraPro~\citep{muller_large-scale_2018}, a large contrastive set of ambiguous pronouns whose antecedents belong to context. In order to validate the improvements achieved by our approaches on the test sets, we perform statistical significance tests, detailed in Annex~\ref{app:significance}.

We experiment with two models: 1) \textbf{base}: a context-agnostic baseline following \textit{Transformer-base} \cite{vaswani_attention_2017}; 2) \textbf{s4to4}: a context-aware concatenation approach with the exact same architecture as \textit{base}, but that adopts sliding windows of 4 concatenated sentences as source and target. An implementation of these models and the proposed approach can be found on github.\footnote{ \href{https://github.com/lorelupo/focused-concat}{https://github.com/lorelupo/focused-concat}}

\begin{table*}[]
\centering
\scalebox{0.9}{
\begin{tabular}{lllll:lcc}
\toprule
En$\rightarrow$Ru system & Deixis & Lex co. & Ell. inf & Ell. vp & Disc. & BLEU & COMET \\
\midrule
base & 50.00 & 45.87 & 51.80 & 27.00 & 46.64 & 31.98 & 0.321 \\
s4to4 & 85.80 & 46.13 & 79.60 & 73.20 & 72.02 & 32.45 & 0.329 \\
s4to4 + \texttt{CD} & \textbf{87.16*} & 46.40 & 81.00 & 78.20* & 73.42* & 32.37 & 0.328 \\
s4to4 + shift + \texttt{CD} & 85.76 & \textbf{48.33*} & \textbf{81.40} & \textbf{80.40*} & \textbf{73.55*} & 32.37 & 0.334* \\
\bottomrule
\end{tabular}
}
\caption{Accuracy on the En$\rightarrow$Ru contrastive set for the evaluation of discourse phenomena (Disc., \%), and BLEU score on the corresponding test set. The accuracy on Disc. is detailed on its left with the accuracy on each of the 4 discourse phenomena evaluated in the contrastive set. The symbol * denotes statistically significant (p < 0.05) improvements w.r.t. base and s4to4.} \label{tab:main_ru}
\end{table*}

\begin{table*}[]
\centering
\scalebox{0.9}{
\begin{tabular}{lllll:lcc}
\toprule
En$\rightarrow$De system  & $d = 1$ & $d = 2$\phantom{.} & $d = 3$\phantom{*} & $d > 3$\phantom{*} & Disc. & BLEU & COMET \\
\midrule
base & 32.89 & 43.97 & 47.99 & 70.58 & 37.27 & 29.63 & 0.546 \\
s4to4 & 68.89 & 74.96 & 79.58 & \textbf{87.78} & 71.35 & 29.48 & 0.536 \\
s4to4 + \texttt{CD} & \textbf{72.86*} & 75.96 & 80.10 & 84.38 & 74.31* & 29.32 & 0.522 \\
s4to4 + shift + \texttt{CD} & 72.56* & \textbf{77.15*} & \textbf{80.27} & 86.65 & \textbf{74.39*} & 29.20 & 0.528 \\
\bottomrule
\end{tabular}
}
\caption{Accuracy on the En$\rightarrow$De contrastive sets for the evaluation of discourse phenomena (Disc., \%), and BLEU score on the corresponding test sets. The accuracy on Disc. is detailed on its left with the accuracy on anaphoric pronouns with antecedents at different distances $d=1,2,...$ (in number of sentences). The symbol * denotes statistically significant (p < 0.05) improvements w.r.t. base and s4to4.} \label{tab:main_de}
\end{table*}

\subsection{Preliminary analysis}\label{subsec:preliminary}

\begin{figure}[]
    \centering
    \includegraphics[width=0.5\textwidth]{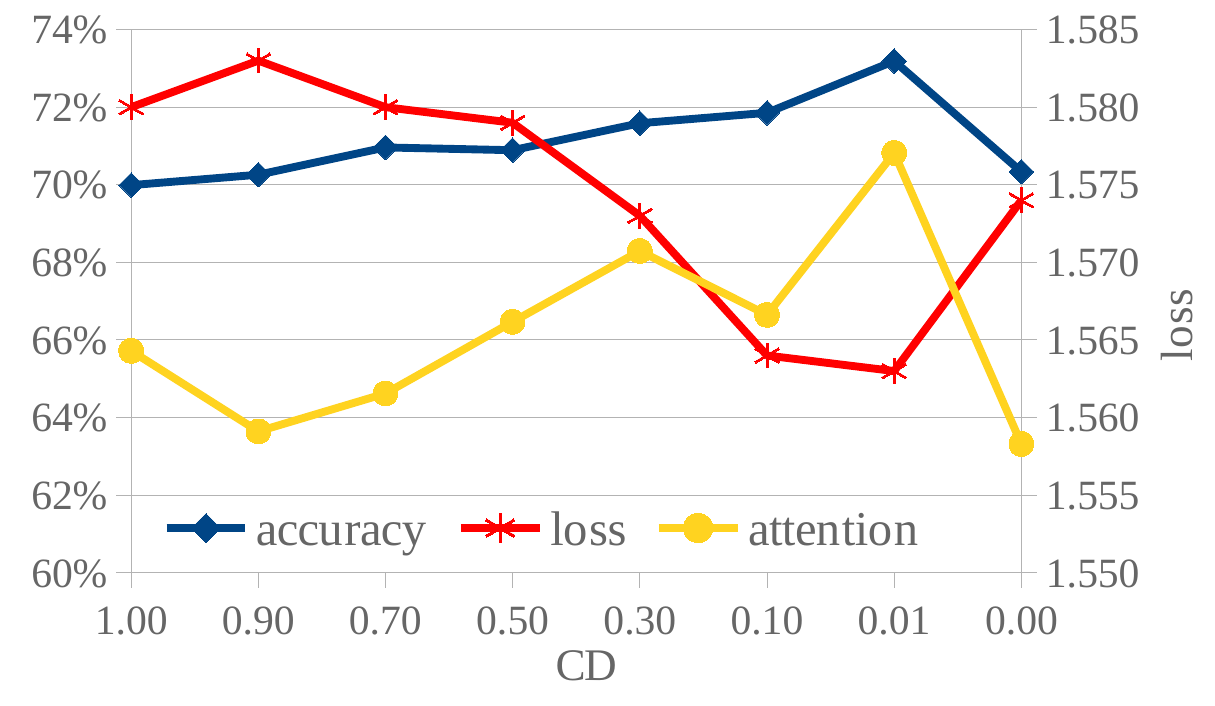}
    \caption{Evaluation of En$\rightarrow$Ru s4to4 trained with various levels of context discounting, ranging from 1 to 0. We plot the best \textit{current loss} obtained by each model on the development set (red), and its average accuracy on the development portion of the contrastive set on discourse phenomena (blue). In yellow, the average portion of attention that is focused on the current sentence (see Section~\ref{subsubsec:cdattn}).}
    \label{fig:best_cd}
\end{figure}

As a preliminary analysis, we evaluate the impact of various values of context discounting on the performance of concatenation approaches with sliding windows, in order to choose one value for all the subsequent experiments. We train En$\rightarrow$Ru s4to4 models with context discounts ranging from 1 (no context discounting) to 0 (context loss is completely ignored): $\texttt{CD}={1.0, 0.9, 0.7, 0.5, 0.3, 0.1, 0.01, 0}$. We evaluate these models on the development sets by means of their average loss calculated over the current target sentence (\textit{current loss}) and the average accuracy on the disambiguation of discourse phenomena. The results are plotted on Figure~\ref{fig:best_cd}. We find out that the stronger the context discounting, the better the performance, with an improving trend from $\texttt{CD}=1$ to $\texttt{CD}=0.01$. Performance drops on the extreme case of $\texttt{CD}=0$, likely because too much training signal is lost in this situation (all the training signal coming from the context is completely ignored). As such, we set $\texttt{CD}=0.01$ for all of our following experiments.

\subsection{Main results}\label{subsec:main}

Tables \ref{tab:main_ru} and \ref{tab:main_de} display the main evaluation results measured in terms of accuracy on contrastive test sets (Disc.) and BLEU, for the En$\rightarrow$Ru and En$\rightarrow$De language pairs, respectively. We first observe that s4to4 is a strong context-aware baseline as it improves accuracy on contrastive sets by a large margin compared to the context-agnostic \textit{base}, as already reported by previous works~\citep{voita_when_2019,zhang_long-short_2020,lopes_document-level_2020}.

Average translation quality as measured by BLEU is virtually the same for all models. Indeed, our main focus is on contrastive evaluation of discourse translation, since average translation quality metrics like BLEU have been repeatedly shown to be ill-equipped to detect improvements in CANMT \cite{hardmeier_discourse_2012}. Learned average translation quality metrics like COMET might be more sensitive to inter-sentential discourse phenomena when applied at document-level, as we do. However, COMET differences are also negligible: all models perform on par according to statistical significance tests, except for the En$\rightarrow$Ru model with context discount and segment shifting, that outperforms all the others with statistical significance.

When evaluating the accuracy on inter-sentential discourse phenomena, instead, we remark relevant performance improvements. In fact, adding a $0.01$ context discounting (+ \texttt{CD}) improves the accuracy on all of the 4 discourse phenomena under evaluation in En$\rightarrow$Ru, and for all distances of pronoun's antecedents in En$\rightarrow$De, with the sole exception of $d>3$, proving to be an effective solution. Adding segment-shifted positions further improves performance for 3 discourse phenomena out of 4, and for pronouns with antecedents at distances $d=1,2$, showing that sliding windows systems often benefit from enhanced sentence position information in order to achieve the discounted CANMT objective. For both language pairs, we adopt a segment-shifting equal to the average sentence length, calculated over the entire training corpus, i.e., +8 positions for En$\rightarrow$Ru and +21 positions for En$\rightarrow$De. Experiments with other shifting values are reported in Section~\ref{subsubsec:shift}.

As a further experiment, we apply our solutions to concatenation models with concatenated windows shorter than 4 sentences,\footnote{We cannot evaluate with more sentences because 4 is the maximum size of documents in the test sets specialized on discourse phenomena.} and evaluate them in the En$\rightarrow$Ru setting. The results presented in Table~\ref{tab:robustness} show that context discounting is effective for s2to2 and s3to3 too, while adding segment-shifted positions only helps s2to2 + \texttt{CD}. As in the case of s4to4, BLEU only displays negligible fluctuations. \\

\begin{table}[]
\centering
\scalebox{0.9}{
\begin{tabular}{lcc}
\toprule
System & Disc.\phantom{*} & BLEU \\
\midrule
s2to2 & 59.10\phantom{*} & 32.73 \\
s2to2 + \texttt{CD} & 60.28* & 32.69 \\
s2to2 + shift + \texttt{CD} & \textbf{60.54*} & 32.41 \\
\midrule
s3to3 & 65.58\phantom{*} & 32.34 \\
s3to3 + \texttt{CD} & \textbf{67.02*} & 32.42 \\
s3to3 + shift + \texttt{CD} & 66.98* & 32.45 \\
\bottomrule
\end{tabular}
}
\caption{Accuracy on the En$\rightarrow$Ru contrastive set for the evaluation of discourse phenomena (Disc., \%), and BLEU score on the test set. The symbol * denotes statistically significant (p < 0.05) improvements w.r.t. s2to2/s3to3. Our approach is effective for different concatenation windows.}\label{tab:robustness}
\end{table}

\subsection{Benchmarking}\label{subsec:benchmarking}

\begin{table*}[]
\centering
\scalebox{0.83}{
\begin{tabular}{lcccc:c|cccc:c}
\toprule
System & \multicolumn{5}{c}{En$\rightarrow$Ru} & \multicolumn{5}{c}{En$\rightarrow$De} \\
 & Deixis & Lex co. & Ell. inf & Ell. vp & Disc. & d=1 & d=2 & d=3 & d>3 & Disc. \\
 \midrule
\citet{chen_breaking_2021} & 62.30 & 47.90 & 64.90 & 36.00 & 55.61 & n.a. & n.a. & n.a. & n.a. & n.a. \\
\citet{sun_rethinking_2022} & 64.70 & 46.30 & 65.90 & 53.00 & 58.13 & n.a. & n.a. & n.a. & n.a. & n.a. \\
\citet{zheng_towards_2020} & 61.30 & 58.10 & 72.20 & 80.00 & 63.30 & n.a. & n.a. & n.a. & n.a. & n.a. \\
\citet{kang_dynamic_2020} & 79.20 & 62.00 & 71.80 & 80.80 & 73.46 & n.a. & n.a. & n.a. & n.a. & n.a. \\
\citet{zhang_long-short_2020} & \textbf{91.00} & 46.90 & 78.20 & \textbf{82.20} & \textbf{75.61} & n.a. & n.a. & n.a. & n.a. & n.a. \\
\hdashline
\citet{maruf_selective_2019} & n.a. & n.a. & n.a. & n.a. & n.a. & 34.70 & 46.40 & 51.10 & 70.10 & 39.15 \\
\citet{voita_context-aware_2018} & n.a. & n.a. & n.a. & n.a. & n.a. & 39.00 & 48.00 & 54.00 & 66.00 & 42.55 \\
\citet{stojanovski_improving_2019} & n.a. & n.a. & n.a. & n.a. & n.a. & 53.00 & 46.00 & 50.00 & 71.00 & 52.55 \\
\citet{lupo_divide_2022} & n.a. & n.a. & n.a. & n.a. & n.a. & 56.50 & 44.90 & 48.70 & 73.30 & 54.98 \\
\citet{muller_large-scale_2018} & n.a. & n.a. & n.a. & n.a. & n.a. & 58.00 & 55.00 & 55.00 & 75.00 & 58.13 \\
s4to4 + shift + \texttt{CD} (ours) & 85.76 & \textbf{48.33} & \textbf{81.40} & 80.40 & 73.56 & \textbf{72.56} & \textbf{77.15} & \textbf{80.27} & \textbf{86.65} & \textbf{74.39} \\
\bottomrule
\end{tabular}
}
\caption{Benchmarking: accuracy (\%) on the contrastive sets for the evaluation of discourse phenomena (Disc., \%).}
\label{tab:sota}
\end{table*}

For a wider contextualization of our results, we compare in Table \ref{tab:sota} our best system with other CANMT systems from the literature. For the En$\rightarrow$Ru language pair, we compare with all the systems from the literature that were trained and evaluated under the same experimental conditions as ours, to the best of our knowledge. In particular, we report the results by \citet{chen_breaking_2021},  \citet{sun_rethinking_2022}' \textit{MR Doc2Doc}, \citet{zheng_towards_2020}, \citet{kang_dynamic_2020}'s \textit{CADec + DCS-pf} and \citet{zhang_long-short_2020}. All of them are sophisticated CANMT systems that add extra trainable parameters to the Transformer architecture. Despite being the simplest and the only parameter free approach, our method outperforms all the others on lexical cohesion and noun phrase inflection based on elided context, while it is only second to \citet{zhang_long-short_2020} on deixis and verb-phrase ellipsis. BLEU scores were not available for comparison on the same test set, except for \citet{zhang_long-short_2020}, which scored 31.84 BLEU points against the 32.45 BLEU points of our method.

For the En$\rightarrow$De language pair, we compare to the literature performing evaluation on \citet{muller_large-scale_2018}'s test set and providing details about their accuracy on pronouns with antecedents at $d>1$. In particular: \citet{maruf_selective_2019}'s best offline system, \citet{stojanovski_improving_2019}'s \textit{pron-25$\rightarrow$pron-0*}, \citet{lupo_divide_2022}'s \textit{K1-d\&r}, \citet{muller_large-scale_2018}'s \textit{s-hier-to-2.tied} and their evaluation of \citet{voita_context-aware_2018}'s architecture.\footnote{Whenever the cited works present and evaluate multiple systems, we compare to the best performing one. To the best of our knowledge, we are including all the relevant works available in the literature. BLEU scores are not compared because, besides using different training data, the cited works don't adopt the same test set neither, with the sole exception of \cite{lupo_divide_2022}.} All of these works but \citet{maruf_selective_2019} adopt the much larger WMT17\footnote{http://www.statmt.org/wmt17/translation-task.html} dataset for training. Despite this advantage, our system outperforms each of them on all the discourse phenomena under evaluation, by a large margin.

Notably, from this comparison it might seem that our approach is proposed in opposition to the others reported in Table \ref{tab:sota}, but it can actually be complimentary to many of them, such as \cite{zhang_long-short_2020}'s, hopefully in a synergistic way. We encourage future research to investigate this possibility.  


\subsection{Analysis of context-discounting}\label{subsec:cdanalysis}

\subsubsection{Loss distribution}\label{subsubsec:loss}

\begin{figure*}[!htb]
    \centering
    \includegraphics[width=1.0\linewidth]{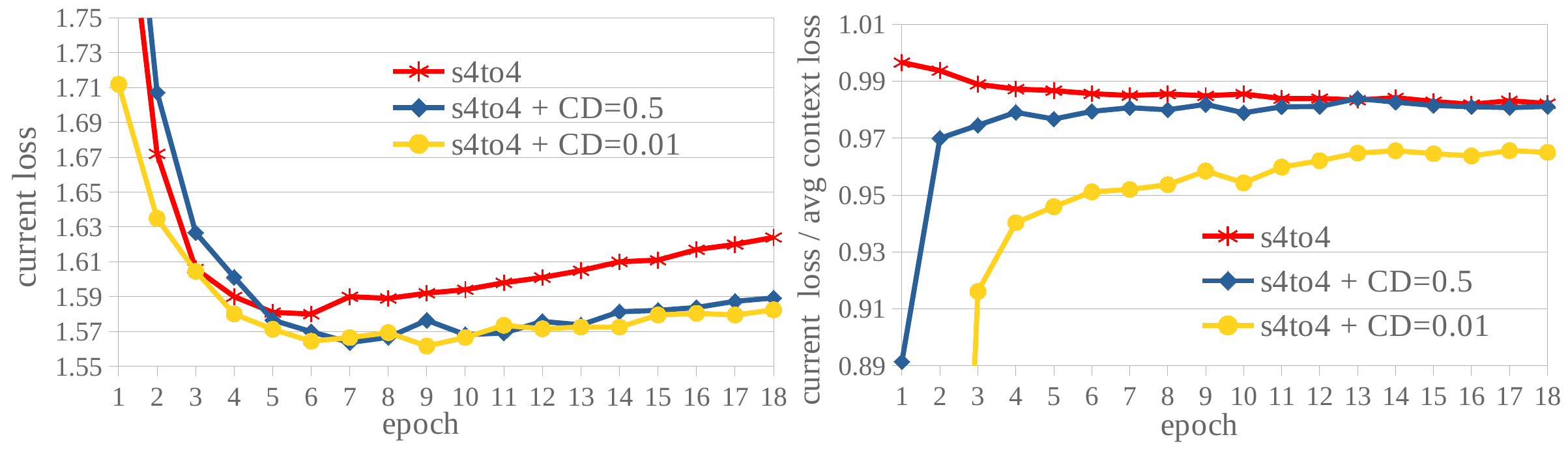}
   \caption{Context discounting enables better predictions of the current sentence (lower validation loss, on the left) at the expense of context sentences (lower current/context validation loss ratio, on the right). Language pair: En$\rightarrow$Ru.}\label{fig:loss}
\end{figure*}

In this section, we analyze the impact of context discounting on the ability of the model to predict the translation of the current sentence. On the left side of Figure~\ref{fig:loss} we plotted the evolution along  training epochs of the loss calculated on the current target sentence (\textit{current loss}), for the En$\rightarrow$Ru language pair. The right side, instead, represents the ratio between the \textit{current loss} and the average loss-per-sentence calculated on the context sentences belonging to the same sliding window. These results support empirically our idea of context discounting as a solution to improve model performance on the current sentence. They also confirm that a strong discounting works best. Interestingly, predictions are improved on the current sentence (left) partially as a result of a trade-off with context quality (right). In fact, the current/context loss ratio of context-discounted models increases along training even when the \textit{current loss} is decreasing, indicating that, at the beginning of training, context discounting pushes the model to only care about current predictions, but later it allows for good predictions of the context too. Such behavior is in line with the intuition that a good translation of the current sentence, even if strongly prioritized, also requires a good translation of the context. Otherwise, it is not possible to systematically solve the translation ambiguities referring to context.

\subsubsection{Attention distribution}\label{subsubsec:cdattn}

In this section we show some empirical evidence in favor of our intuition that context-discounting improves performance by helping the self-attentive mechanism to be more focused on the current sentence (less distracted by context). We analyzed the distribution of the self-attention weights generated by the queries belonging to the current sentence (\textit{current queries}), and how it is impacted by context discounting. Figure~\ref{fig:best_cd} clearly shows that context-discounting impacts the distribution of attention weights by skewing it towards the current sentence: a higher percentage of the total attention from \textit{current queries} is directed towards tokens belonging to the (same) current sentence.
As expected, the higher the context-discounting, the higher the portion of attention that is not dispersed towards context. The limit case of $CD = 0$ is not aligned with this trend, however. We suspect that the attention distribution is more flat in this case because the model encounters learning difficulties due to the training signal from the context being completely ignored (c.f. \citet{bao_g-transformer_2021} on non-fully-converged models having a flatter attention distribution).   \\ 

\subsubsection{Robustness}\label{subsubsec:robustness}

\begin{figure}[t]
    \centering
    \includegraphics[width=0.52\textwidth]{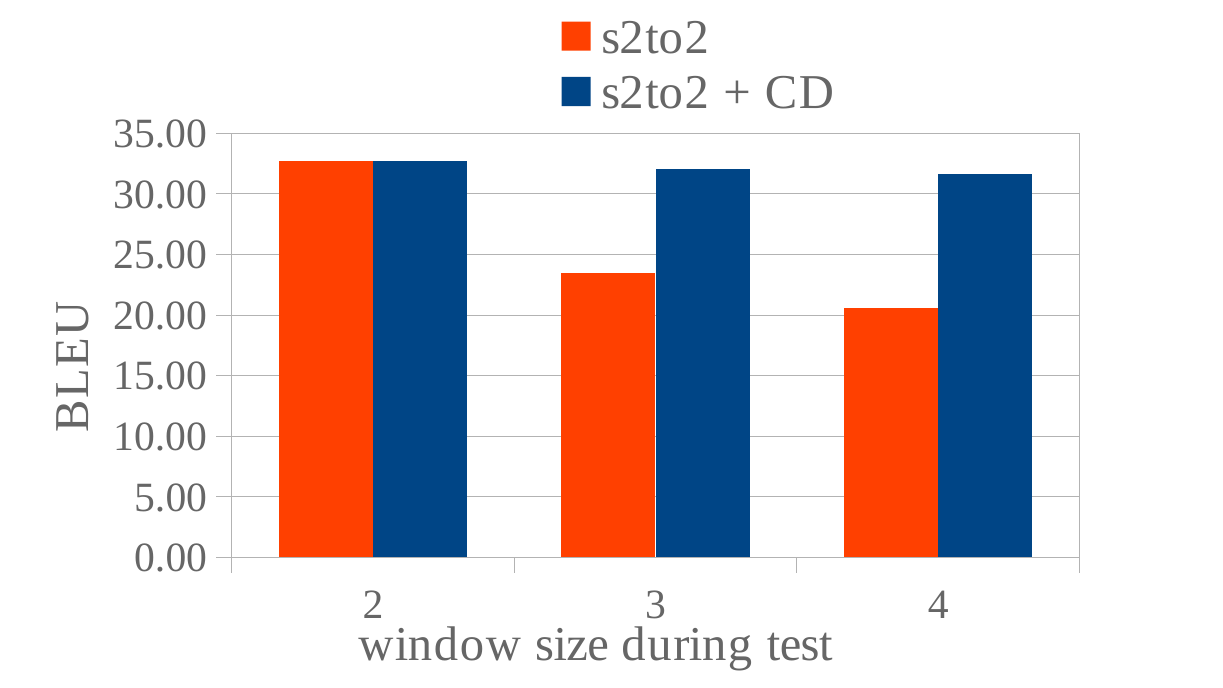}
    \caption{Our approach improves robustness of En$\rightarrow$Ru s2to2 to window sizes unseen during training.}\label{fig:robustness}
\end{figure}

Figure~\ref{fig:robustness} shows that the s2to2 model is not robust to the translation of concatenation windows longer than those seen during training, i.e. longer than 2 sentences. Indeed, s2to2 loses $9.23$ BLEU points when translating the same test set with windows of 3 sentences, and $12.14$ BLEU points when translating with windows of 4. Instead, the context discounted model (blue bars) is very robust to unseen context lengths, being capable of translating them with minor degradation in average translation quality ($-0.68$ and $-1.06$ BLEU points for windows of 3 and 4, respectively). We observe a similar trend for s3to3, that loses $1.74$ BLEU points when tested with windows of size 4, but recovers completely when equipped with context-discounting. The increased robustness of the concatenation models w.r.t. context size suggests once again that context discounting helps the models focusing on the current sentence.

\subsection{Analysis of segment-shifted positions}\label{subsec:seganalysis}

\subsubsection{Attention distribution}\label{subsubsec:entropy}

As a complementary evaluation, we tested if segment-shifted positions work as intended, i.e., by helping context-discounted models to learn the locality properties of both the language itself and the source-target alignment~\cite{hardmeier_discourse_2012,rizzi_locality_2013}. In other words, we expect segment-shifted positions to result in a more localized attention-distribution, in each of the sentences belonging to the concatenated sequence. To this aim, we computed the average entropy of the distribution of attention weights generated by all queries (both from current and context sentences), in both self and cross-attention. Results are shown in Table~\ref{tab:entropy}: context-discounting slightly reduces the average entropy, and this effect is amplified with the adoption of segment-shifted positions. Segment-shifted  positions make attention more focused locally, as intended, which explains why the job of context discounting is eased by this solution.

\begin{table}[]
\centering
\scalebox{0.9}{
\begin{tabular}{lc}
\toprule
System & Attn entropy \\
\midrule
s4to4 & 2.293 \\
s4to4 + \texttt{CD} & 2.276 \\
s4to4 + shift + \texttt{CD} & \textbf{2.251} \\
\bottomrule
\end{tabular}
}
\caption{Average entropy of self and cross-attention weights decreases with the help of context-discounting and segment-shifted positions. All of the three values are different from one another with statistical significance (p<0.01).}\label{tab:entropy}
\end{table}

\subsubsection{Comparison with segment-embeddings}\label{subsubsec:segemb}

In this section we compare our parameter-free approach to include explicit information on segment position (segment-shifted positions), with learned segment embeddings \cite{devlin_bert_2019}, and sinusoidal segment embeddings. The latter are added to token and position embeddings at input, in the very same way as learned segment embeddings, with the only difference that their parameters are not learned but defined in the same way as sinusoidal position embeddings \cite{vaswani_attention_2017}. In order to evaluate which approach helps best with context-discounting, we trained a context-discounted concatenation model with learned segment embeddings (s4to4+lrn+\texttt{CD}), and one with sinusoidal segment embeddings (s4to4+sin+\texttt{CD}), and compared them with s4to4+shift+\texttt{CD}. The results reported in Table~\ref{tab:segemb} do not display any statistically significant differences across the three alternatives (p>0.05), except for learned embeddings, that underperform with statistical significance the other two variants on En$\rightarrow$De. Instead, sinusoidal segment embeddings are competitive with segment-shifted positions on both language pairs.
We leave a more in-depth analysis of segment-embeddings for concatenation approaches to future works.

\begin{table}[]
\small
\centering
\begin{tabular}{lcccc}
\toprule
 & \multicolumn{2}{c}{En$\rightarrow$Ru} & \multicolumn{2}{c}{En$\rightarrow$De} \\
System & Disc. & BLEU & Disc. & BLEU \\
\midrule
s4to4 + shift + \texttt{CD} & 73.56 & 32.45 & \textbf{74.39} & 29.20 \\
s4to4 + lrn + \texttt{CD} & \textbf{73.68} & 32.45 & 72.14 & 28.35 \\
s4to4 + sin + \texttt{CD} & 73.48 & 32.53 & 73.88 & 29.23 \\
\bottomrule
\end{tabular}
\caption{Comparison between segment-shifted positions, learned segment embeddings and sinusoidal segment embeddings. Approaches are evaluated with accuracy on contrastive sets for the evaluation of discourse phenomena (Disc., \%), and BLEU score on test sets. Differences across models are not statistically significant (p>0.05), except for s4to4+lrn+\texttt{CD} on En$\rightarrow$De.}\label{tab:segemb}
\end{table}

\subsubsection{Segment-shifting variants}\label{subsubsec:shift}

\begin{table}[]
\centering
\small
\begin{tabular}{cccc}
\toprule
System & Shift & Disc. & BLEU \\
\midrule
s4to4 + shift + \texttt{CD} & 100.00 & 73.46 & 32.41 \\
s4to4 + shift + \texttt{CD} & avg-sequence & \textbf{73.86} & 32.37 \\
s4to4 + shift + \texttt{CD} & avg-corpus & 73.56 & 32.45 \\
\bottomrule
\end{tabular}
\caption{Accuracy on the En$\rightarrow$Ru contrastive set for the evaluation of discourse phenomena (Disc., \%), and BLEU score on the test set. Differences across models are not statistically significant (p>0.05).}\label{tab:shift}
\end{table}

In the experiments reported above, we always adopt a shifting value equal to the average sentence length calculated over the entire training corpus (avg-corpus), i.e., +8 positions for En$\rightarrow$Ru, +21 positions for En$\rightarrow$De. In this section we evaluate two alternative strategies for the selection of the shifting value: 1) applying a big shift of 100 units, one order of magnitude bigger than the average sentence length in the corpus (100); 2) applying a shifting value equal to the average sentence length of each window, calculated dynamically for each window of 4 concatenated sentences (avg-sequence). The results of this study are reported in Table~\ref{tab:shift}. We do not observe relevant differences in average translation quality (BLEU) nor accuracy on the translation of discourse phenomena, and therefore confirm that the avg-corpus approach is a good alternative.

\section{Conclusions}
We presented a simple, parameter-free modification of the NMT objective for context-aware translation with sliding windows of concatenated sentences: context discounting. We analyzed the impact of our approach in the trade-off between current sentence predictions and context sentence predictions, showing that context discounting helps the model to focus on the current sentence, as intended. As a result, the concatenation model significantly improves its ability to disambiguate inter-sentential discourse phenomena, and becomes more robust to different context sizes. As an additional inductive bias towards locality, we equipped our model with segment-shifted positions, marking more explicitly the boundaries between sentences. This solution brings further improvements on targeted evaluation metrics. In the attempt of explaining the empirical functioning of the proposed solutions, we analysed their impact on the distribution of the attention weights, showing that they make it more focused and skewed towards the current sentence, as intended.

\section*{Limitations and future works}
Our experiments are limited to the use case of short concatenated windows (up to 4 sentences). This is enough for capturing most of the ambiguous inter-sentential discourse phenomena, that usually span across a few sentences only~\cite{muller_large-scale_2018,voita_when_2019,lupo_divide_2022}.  However, recent works suggest that longer context windows might be helpful to increase average translation quality (BLEU) of concatenation approaches~\cite{junczys-dowmunt_microsoft_2019,bao_g-transformer_2021,sun_rethinking_2022}, and long-range discourse phenomena could be handled. We hope to investigate the impact of context discounting on longer sequences in future works. We also encourage to test the effectiveness of our approach on a wider range of data scenarios: from very limited document-level data to very abundant, including back translation \cite{ma_comparison_2021} and monolingual pre-training techniques~\cite{junczys-dowmunt_microsoft_2019,sun_rethinking_2022}, to understand whether these methods are only alternative to context discounting or there exist synergies.  Furthermore, experimenting with future context is also needed (c.f.~\citet{wong_contextual_2020}).

\section*{Acknowledgements}
We thank the anonymous reviewers for their insightful comments.
This work has been partially supported by the Multidisciplinary Institute in Artificial Intelligence MIAI@Grenoble Alpes (ANR-19-P3IA-0003).

\bibliography{document_nmt}
\bibliographystyle{acl_natbib}

\appendix

\section{Details on experimental setup}\label{app:setup}
All models are implemented in \textit{fairseq}~\cite{ott_fairseq_2019} and follow the \textit{Transformer-base} architecture \cite{vaswani_attention_2017}:  hidden size of 512, feed forward size of 2048,  6  layers,  8  attention  heads, total 60.7M parameters. They are trained on 4 Tesla V100, with a fixed batch size of approximately 32k tokens for En$\rightarrow$Ru and 16k for En$\rightarrow$De. As it has been shown that Transformers need a large batch size for achieving the best performance \citep{popel_training_2018}. We stop training after 12 consecutive non-improving validation steps (in terms of loss on dev), and we average the weights of the 5 checkpoints that are closest to the best performing checkpoint, included. We train models with the optimizer configuration and learning  rate (LR)  schedule  described in \citet{vaswani_attention_2017}. The maximum LR is optimized for each model over the search space $\{7e-4,9e-4,1e-3,3e-3\}$. The LR achieving the best loss on the validation set after convergence was selected. We use label smoothing with an epsilon value of 0.1~\citep{Pereyra_regularizing_2017} for all settings. We adopt strong model regularization (dropout=0.3) following~\citet{kim_when_2019} and \citet{ma_comparison_2021}. At inference time, we use beam search with a beam of 4 for all models. We adopt a length penalty 0.6 for all models. The other hyperparameters were set according to the relevant literature~\citep{vaswani_attention_2017, popel_training_2018, voita_when_2019, ma_comparison_2021, lopes_document-level_2020}.

\subsection{Statistical hypothesis tests}\label{app:significance}

We perform statistical hypothesis testing with McNemar's test \citet{mcnemar_note_1947} for comparing accuracy results on the contrastive test sets. For comparing BLEU performances and mean entropy (Table~\ref{tab:entropy}), we use approximate randomization \cite{riezler_pitfalls_2005} with 10000 and 1000 permutations, respectively. For COMET, the official library\footnote{https://github.com/Unbabel/COMET} has a built in tool for the calculation of statistical significance with Paired T-Test and bootstrap resampling \cite{koehn_statistical_2004}.

\section{Details on experimental results}\label{app:results}

In this section, we report more details about the results presented in our Tables.

\subsection{Evaluation of the translation of discourse phenomena}\label{app:discourse}

For each model that we evaluated by its accuracy on the contrastive sets for the evaluation of discourse phenomena (Disc., \%), we include in Table~\ref{tab:disc} the accuracy achieved on the different subsets of the contrastive sets, as already done for Tables \ref{tab:main_ru}, \ref{tab:main_de} and \ref{tab:sota}. For the En$\rightarrow$Ru set \cite{voita_when_2019}, we report the accuracy on each of the 4 discourse phenomena under evaluation; for the En$\rightarrow$De test set \cite{muller_large-scale_2018}, the accuracy on anaphoric pronouns with antecedents at different distances $d=1,2,...$ (in number of sentences). As it can be noticed, our approach mostly outperform baselines and other variants on the majority of the evaluation subsets. We also include the column Disc$_{\text{avg}}$, which is calculated, for both language pairs, as the average of the 4 columns before the vertical dashed line.

{\tiny
\begin{align*}
Disc. &= \frac{d1*7075+d2*1510+d3*573+(d>3)*442}{9600}, \\
Disc_{\text{avg}} &=\frac{d1+d2+d3+d>3}{4}.
\end{align*}
}%

Disc$_{\text{avg}}$ represents the average accuracy on the disambiguation of the discourse phenomena present in the contrastive sets, as if they were all present with the same frequency. Instead, Disc. represents the overall accuracy on the contrastive set, which is equivalent to the average over the same 4 columns, but weighted by the sample size (last row) of each penomenon represented by the columns. While Disc. is a proxy of the ability to correctly translate a distribution of inter-sentential discourse phenomena as represented in the contrastive set, Disc$_{\text{avg}}$ is a proxy for the average ability to translate each of the inter-sentential phenomena under evaluation. Interestingly, Disc$_{\text{avg}}$ captures more evidently than Disc. the improvement achieved by adding segment-shifted positions to the context-discounted concatenation models. Finally, Disc$_{all-d}$ is calculated like Disc. but it also take into account pronouns whose antecedent belong to the same sentence ($d=0$, i.e., they don't require context).

\begin{table*}[]
\centering
\scalebox{0.63}{
\begin{tabular}{lcccc:cc|ccccc:ccc}
\toprule
 & \multicolumn{6}{c}{En$\rightarrow$Ru} &  &  & \multicolumn{6}{c}{En$\rightarrow$De} \\
 \midrule
System & Deixis & Lex co. & Ell. inf & Ell. vp & Disc. & Disc$_{\text{avg}}$ & d=0 & d=1 & d=2 & d=3 & d>3 & Disc. & Disc$_{\text{avg}}$ & Disc$_{all-d}$ \\
base & 50.00 & 45.87 & 51.80 & 27.00 & 46.64 & 47.67 & 68.75 & 32.89 & 43.97 & 47.99 & 70.58 & 37.27 & 48.86 & 43.57 \\
s4to4 & 85.80 & 46.13 & 79.60 & 73.20 & 72.02 & 71.18 & 75.20 & 68.89 & 74.96 & 79.58 & \textbf{87.78} & 71.35 & 77.80 & 72.12 \\
s4to4 + \texttt{CD} & \textbf{87.16} & 46.40 & 81.00 & 78.20 & 73.42 & 73.19 & \textbf{76.66} & \textbf{72.86} & 75.96 & 80.10 & 84.38 & 74.31 & 78.33 & \textbf{74.78} \\
s4to4 + shift + \texttt{CD} & 85.76 & \textbf{48.33} & \textbf{81.40} & \textbf{80.40} & \textbf{73.56} & \textbf{73.97} & 75.25 & 72.56 & \textbf{77.15} & \textbf{80.27} & 86.65 & \textbf{74.39} & \textbf{79.16} & 74.56 \\
\hdashline
s4to4 + sin + \texttt{CD} & 87.96 & \textbf{46.80} & 78.00 & \textbf{76.60} & 73.48 & 72.34 & \textbf{76.75} & \textbf{71.83} & \textbf{76.82} & \textbf{80.97} & \textbf{87.55} & \textbf{73.88} & \textbf{79.29} & \textbf{74.46} \\
s4to4 + lrn + \texttt{CD} & \textbf{88.12} & 46.47 & \textbf{81.20} & 75.60 & \textbf{73.68} & \textbf{72.85} & 73.91 & 70.21 & 75.29 & 77.66 & 85.06 & 72.14 & 77.06 & 72.49 \\
\hdashline
s4to4 + 100 + \texttt{CD} & \textbf{85.60} & \textbf{48.73} & \textbf{80.80} & \textbf{79.60} & \textbf{73.46} & \textbf{73.68} & n.a. & n.a. & n.a. & n.a. & n.a. & n.a. & n.a. & n.a. \\
s4to4 + avg-seq + \texttt{CD} & 84.84 & 46.20 & 77.60 & 73.00 & 71.34 & 70.41 & n.a. & n.a. & n.a. & n.a. & n.a. & n.a. & n.a. & n.a. \\
\hdashline
s2to2 & 61.84 & 46.07 & 74.60 & 69.00 & 59.10 & 62.88 & n.a. & n.a. & n.a. & n.a. & n.a. & n.a. & n.a. & n.a. \\
s2to2 + \texttt{CD} & \textbf{62.88} & 46.27 & 78.00 & \textbf{71.60} & 60.28 & 64.69 & n.a. & n.a. & n.a. & n.a. & n.a. & n.a. & n.a. & n.a. \\
s2to2 + shift + \texttt{CD} & 62.60 & \textbf{46.60} & \textbf{81.20} & 71.40 & \textbf{60.54} & \textbf{65.45} & n.a. & n.a. & n.a. & n.a. & n.a. & n.a. & n.a. & n.a. \\
\hdashline
s3to3 & 73.52 & 45.87 & 78.00 & 72.60 & 65.58 & 66.45 & n.a. & n.a. & n.a. & n.a. & n.a. & n.a. & n.a. & n.a. \\
s3to3 + \texttt{CD} & 73.88 & \textbf{46.80} & \textbf{82.40} & \textbf{78.00} & \textbf{67.02} & \textbf{67.45} & n.a. & n.a. & n.a. & n.a. & n.a. & n.a. & n.a. & n.a. \\
s3to3 + shift + \texttt{CD} & \textbf{75.24} & 46.07 & 79.40 & 76.00 & 66.98 & 68.45 & n.a. & n.a. & n.a. & n.a. & n.a. & n.a. & n.a. & n.a. \\
\hdashline
\citet{chen_breaking_2021} & 62.30 & 47.90 & 64.90 & 36.00 & 55.61 & 52.78 & n.a. & n.a. & n.a. & n.a. & n.a. & n.a. & n.a. & n.a. \\
\citet{sun_rethinking_2022} & 64.70 & 46.30 & 65.90 & 53.00 & 58.13 & 57.48 & n.a. & n.a. & n.a. & n.a. & n.a. & n.a. & n.a. & n.a. \\
\citet{zheng_towards_2020} & 61.30 & 58.10 & 72.20 & 80.00 & 63.30 & 67.90 & n.a. & n.a. & n.a. & n.a. & n.a. & n.a. & n.a. & n.a. \\
\citet{kang_dynamic_2020} & 79.20 & 62.00 & 71.80 & 80.80 & 73.46 & 73.45 & n.a. & n.a. & n.a. & n.a. & n.a. & n.a. & n.a. & n.a. \\
\citet{zhang_long-short_2020} & \textbf{91.00} & 46.90 & 78.20 & \textbf{82.20} & \textbf{75.61} & \textbf{74.58} & n.a. & n.a. & n.a. & n.a. & n.a. & n.a. & n.a. & n.a. \\
\citep{maruf_selective_2019} & n.a. & n.a. & n.a. & n.a. & n.a. & n.a. & 68.60 & 34.70 & 46.40 & 51.10 & 70.10 & 39.15 & 50.58 & 45.04 \\
\citep{muller_large-scale_2018} & n.a. & n.a. & n.a. & n.a. & n.a. & n.a. & 75.00 & 39.00 & 48.00 & 54.00 & 66.00 & 42.55 & 51.75 & 49.04 \\
\citep{stojanovski_improving_2019} & n.a. & n.a. & n.a. & n.a. & n.a. & n.a. & 74.00 & 53.00 & 46.00 & 50.00 & 71.00 & 52.55 & 55.00 & 56.84 \\
\citep{lupo_divide_2022} & n.a. & n.a. & n.a. & n.a. & n.a. & n.a. & \textbf{81.10} & 56.50 & 44.90 & 48.70 & 73.30 & 54.98 & 55.85 & \textbf{60.21} \\
\citep{muller_large-scale_2018} & n.a. & n.a. & n.a. & n.a. & n.a. & n.a. & 65.00 & \textbf{58.00} & \textbf{55.00} & \textbf{55.00} & \textbf{75.00} & \textbf{58.13} & \textbf{60.75} & 59.51 \\
\midrule
Sample size & 2500 & 1500 & 500 & 500 & 5000 & 5000 & 2400 & 7075 & 1510 & 573 & 442 & 9600 & 9600 & 12000 \\
\bottomrule
\end{tabular}
}\caption{Accuracy on contrastive sets for the evaluation of discourse phenomena (Disc., \%) and on their subsets: for En$\rightarrow$Ru, the accuracy on each of the 4 discourse phenomena under evaluation; for En$\rightarrow$De, the accuracy on anaphoric pronouns with antecedents at different distances $d=1,2,...$ (in number of sentences). Disc$_{all-d}$, includes also $d=0$. Disc$_{\text{avg}}$ denotes the average of the 4 accuracies before the dashed line.
}\label{tab:disc}
\end{table*}

\end{document}